\documentclass[a4paper]{article}

\usepackage{INTERSPEECH2022}

\usepackage{hyperref}
\hypersetup{
    colorlinks=true,
    linkcolor=blue,
    filecolor=magenta,      
    urlcolor=cyan,
    pdftitle={Overleaf Example},
    pdfpagemode=FullScreen,
    }

\usepackage{xcolor}

\title{When Is TTS Augmentation Through a Pivot Language Useful?}
\name{Nathaniel Robinson, Perez Ogayo, Swetha Gangu,  David R. Mortensen, Shinji Watanabe}

\address{
  Language Technologies Institute\\
  Carnegie Mellon University\\
  Pittsburgh, PA, USA
 }

\email{\{nrrobins, aogayo, sgangu,   dmortens, swatanab\}@cs.cmu.edu}

\begin{document}

\maketitle
\begin{abstract}

Developing Automatic Speech Recognition (ASR) for low-resource languages is a challenge due to the small amount of transcribed audio data. For many such languages,  audio and text are available separately, but not audio with transcriptions. Using text, speech can be synthetically produced via text-to-speech (TTS) systems. However, many low-resource languages do not have quality TTS systems either. We propose an alternative: produce synthetic audio by running text from the target language through a trained TTS system for a higher-resource pivot language. We investigate when and how this technique is most effective in low-resource settings. In our experiments, using several thousand synthetic TTS text-speech pairs and duplicating authentic data to balance yields optimal results. Our findings suggest that searching over a set of candidate pivot languages can lead to marginal improvements and that, surprisingly, ASR performance can by harmed by increases in measured TTS quality. Application of these findings improves ASR by 64.5\% and  45.0\% character error reduction rate (CERR) respectively for two low-resource languages: Guaraní and Suba.
%We present experiments with this technique in the context of Kiswahili ASR and Guaraní ASR. Our findings suggest that this novel TTS augmentation method is most effective in low-resource ASR settings, but that, surprisingly, outcomes do not depend on TTS quality. In our analysis to remedy this, we show that the most important factor for success in this method is controlling for text domain or genre. We will release all software and datasets upon acceptance.%We show that this novel TTS augmentation depends on the amount of authentic training data and that it is most effective in lowest-resource ASR settings. Our findings also suggest, surprisingly, that orthographic transliteration into the high-resource language prior to TTS does not improve performance. \nrr{Update to include domain experiment \ref{sxn:dom_abl}}

\end{abstract}

\section{Problem statement and motivation}

%\quad
Applications of ASR systems such as digital assistants are becoming increasingly ubiquitous. Despite ASR being a crucial task for low-resource and endangered languages, most existing ASR projects cover high-resource languages and dialects from industrialized nations; most low-resource languages are left behind. This is troubling because speakers of low-resource languages can benefit significantly from ASR. ASR technologies could enable them to access digital information relating to education, politics, health conditions, natural disasters, etc. Endangered languages particularly need ASR for documentation, revitalization, and learning resources, since human audio transcription is prohibitively slow \cite{michaud2014towards,shi-etal-2021-leveraging}.  

Most state-of-the-art ASR architectures for high-resource languages require large data sets, which take extensive time and resources to collect \cite{asrlorec}. This work explores data augmentation for ASR via TTS for low resource languages, with Coastal Kiswahili (SWH), Guaraní (GRN), and Suba (SXB) as case studies, and Italian (ITA) as an additional example to demonstrate training trends. Kiswahili is a Bantu language and lingua franca in East and Central Africa. It is spoken by $\sim$200 million people in Africa \cite{conversation_2022}. Though it is the most spoken African language, few Kiswahili speech technology developments have been made \cite{asrlorec, getao2006creation}. Guaraní is South American  language of the Tupi-Guaraní family  with $\sim$5.85 million speakers in Paraguay and for which digital resources are becoming increasingly necessary, though they remain limited \cite{jenner_2019}.
% \footnote{According to Guaraní expert and linguist David Olivera, Guaraní ought to be moved into the digital sphere in order to revitalize the language further and maintain its relevance \cite{jenner_2019}.}. 
% It was recognized as an official language of Paraguay in 1992 and given equal status with Spanish. 
Suba is an endangered Bantu language with less than 10000 speakers in Kenya \cite{bernadine_racoma_olusuba_2014}, for which speech technologies can help revitalization efforts.
 
Previous research \cite{rossenbach2020generating, 9053139, zheng2021using} shows that TTS data can be used to augment ASR training. However, many low-resource languages that can benefit from ASR data augmentation do not have TTS systems. To this end, we ask: \emph{Can we use synthesized speech from a high-resource language's TTS system to improve ASR in a low-resource language?}
 
This question prompts a few others. When augmenting, how much synthesized audio should be used? Which high-resource language should be used for TTS? In initial experiments to address these questions, we noticed that pivot-language TTS audio is noisy to human perception. This led us to experiment with TTS quality improvement. We contribute:

 \begin{itemize}
   \setlength{\itemsep}{1pt}
  \item A novel study on the effects of TTS augmentation through a pivot language for low-resource ASR, with accompanying software and datasets % and a pipeline for incorpoating TTS augmentation into ESPNet2 \cite{watanabe20202020}.
  \item Experimentally backed recommendations for augmentation parameters: data amount and duplication, choice of pivot language, and TTS quality, with surprising findings suggesting language relatedness and impressionistic TTS quality may not improve performance
  \item ASR improvement across languages, including 64.5\% and 45.0\% CERR\footnote{We use reduction rates calculated as $\frac{rate_{old} - rate_{new}}{rate_{old}}$} for low-resource Guaraní and Suba
%   \item Use TTS systems for our target language to generate data to augment datasets.
%   \item Use TTS systems from other languages to create synthetic data in the absence of a TTS system for the target language. These languages should have similar pronunciations and phones. We examine how the usage of different TTS systems impacts ASR model’s performance. 
\end{itemize}

% We present trends in authentic and TTS-augmented data amounts on ASR performance. And we explore the effects of three additional factors on TTS augmentation efficacy: TTS quality, language similarity, and text domain consistency. We conclude that domain consistency is by far the most important.

\section{ Related work}

% \nrr{Make sure it's emphasized somewhere what is novel exactly. Fix language intro too}

Multiple data augmentation techniques exist for low-resource ASR. Popular approaches include SpecAugment \cite{2019}, language models (LM), and incorporating text and untranscribed speech in addition to traditional waveform variation methods such as PSOLA, time-stretching and noise addition \cite{asrlorec}. 

We are not the first to explore cross-lingual transfer for low-resource ASR. Pre-training weights on speech-to-text translation from a high-resource language can reduce ASR error rates \cite{Wang2020ImprovingCT}, and high-resource ASR pre-training can conversely improve low-resource translation \cite{Stoian2020AnalyzingAP, Bansal2019PretrainingOH}. Our work likewise seeks to improve ASR for languages where text data is more available than transcribed speech, but we do so via TTS, without pre-training or requiring translations or translation systems.

% \cite{8923082} explore an effective approach where transfer learning is performed on a heavily concentrated synthetic dataset, followed by fine-tuning using limited synthetic data. \cite{DBLP:journals/corr/abs-1910-10762} ask questions regarding what factors behind pre-training models with higher-resourced data are most pertinent to boosts in ASR performance for the target language. Specifically, they focus on whether pre-training with a larger number of higher-resourced languages is more important than the overall amount of pre-training data. They conclude that neither of these factors are more important than the other, but the WER of the pre-trained model is the most likely indicator of performance. This seems to be because these models with lower WERs are encoding more language universal phonetic information. 

Researchers have also explored using TTS to augment ASR training. \cite{rossenbach2020generating, 9053139, jasonli2018} explored and implemented speaker augmentation to increase variability for synthesized speech, but \cite{8639589} showed significant improvements in word error rate (WER) from a single speaker. 
\cite{rossenbach2020generating} explored how LM, SpecAugment, and TTS augmentation affect WER. They found that these methods are independent of each other and that using TTS data yielded more improvements than LM and SpecAugment, with the best configuration combining them all. In all these  works, researchers trained neural TTS models in the target language, requiring more than ten hours of high-quality data \cite{8683862}. We build upon and differ from these works: because we use TTS systems trained on high-resource languages, our approach may apply to the thousands of low-resource languages for which ten-hour audio data sets are not available \cite{info11020067}.
% Our approach takes advantage of cross-lingual transfer learning and, in theory, can be applied to a much more comprehensive array of languages (since most of the world's $\sim7000$ languages are low-resource \cite{info11020067}). %We also investigate how data amount and balance, relation between pivot and target languages, and quality of TTS systems affect ASR error rates.

Other researchers have asked how much synthetic TTS data is appropriate, since acoustic differences between TTS data and authentic data can make it less effective from the same text \cite{jasonli2018,fazel2021synthasr}. \cite{jasonli2018} found the best synthetic/authentic data balance on a LibriSpeech task was 50/50. \S \ref{sxn:synth_data_amt} shows similar findings in our novel setting of TTS from a high-resource pivot language.

\section{Approach}

% \nrr{Talk about hyperparams}\pao{Can we just say that we include the configurations in the repo that we release to save on space coz I don't think we focus on them?} To evaluate the systems, we use character error rate (CER) and word error rate (WER).

\begin{figure}[h!]
    \centering
    \includegraphics[width=6cm]{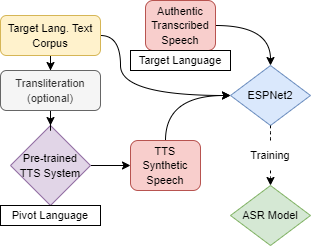}
\caption{\label{pipeline}
        \centering
    Visual depiction of our TTS augmentation pipeline %\nrr{Add optional transliteration step and shrink diagram}
    }
\end{figure}

\subsection{Data setup}
\label{sxn:data_method}
%\quad
\textbf{Authentic data} We used Kiswahili audio with transcriptions from the Gamayun Swahili Minikit\footnote{\url{https://gamayun.translatorswb.org/data}} \cite{9342939}, which contains  4700 transcribed recordings, for training, validation, and testing. %The dataset contains 4700 labeled recordings.
% We used 3900 for training, 400 as a dev set, and 400 for testing. In our low-resource experiments we used only 300 recordings for training, 100 for the dev set, and 100 for testing. \nrr{Need to use the same test set} 
We obtain Italian and Guaraní data from Mozilla Common Voice\footnote{\url{https://commonvoice.mozilla.org/en/datasets}} Corpus 7.0. Our Suba corpus presents an extremely low-resource setting: 1178 sentences (1.7hrs) obtained from AfricanVoices \cite{ogayo_black}.

\textbf{TTS Augmentation} We feed authentic text to TTS models to generate synthetic audio which results into a synthetic text-speech pair. We used Microsoft TTS\footnote{\url{https://azure.microsoft.com/en-us/services/cognitive-services/text-to-speech}} and Google Cloud\footnote{\url{https://cloud.google.com/text-to-speech}} neural models to obtain synthetic data. We employed this augmentation for ASR in four languages: Kiswahili, Guaraní, Italian, and Suba. We outline the text corpora we used as TTS prompts for each language. \textit{Kiswahili}: 14737 Kiswahili sentences from the Helsinki Swahili corpus \cite{hcs-a-v2_en}, which contains news and political text. We selected these sentences to be phonetically diverse using Festvox tools \cite{anumanchipalli2011festvox}. \textit{Guaraní}: a uniformly random mixture from Guaraní Wikipedia\footnote{\url{https://wortschatz.uni-leipzig.de/en/download/Guarani}} and another Guaraní corpus \cite{chiruzzo-etal-2020-development}. \textit{Italian}: text from unused recordings in the Mozilla data we downloaded for training. \textit{Suba}: text from the Suba New Testament \cite{noauthor_suba_2018}.  Further data details are in Table \ref{tab_data}.

% Because this text is a different genre from the Gamayun set, we experiment with using Gamayun transcriptions in its stead in section \ref{sxn:dom_abl}. (This yields the largest improvements over baseline results.)

\textbf{Transliteration}
We found that pivot language TTS systems make many pronunciation errors and hallucinate phones when Kiswahili text is input. This is not surprising because of orthographic diversity across languages (especially in the case of languages that use different alphabets, such as Kiswahili and Arabic). To remedy this in some experiments, we transliterated text into pivot language orthography using hand-crafted phone maps\footnote{We release the exact mappings at \url{https://github.com/n8rob/Multilingual\_TTS\_Augmentation}. Target language phones not present in the pivot language are approximated with close equivalents (e.g. representing Kiswahili's voiced post-alveolar affricate with Spanish's voiceless post-alveolar affricate).}. This generated empirically higher quality audio with TTS systems for those languages. (See \S \ref{sxn:tts_qual}.) This is represented as the optional step "Transliteration" in Figure \ref{pipeline}. 
%We applied this same strategy to Guaraní ASR and to Italian ASR in our results analysis. We transliterated 2000 sentences from Guaraní Wikipedia\footnote{https://wortschatz.uni-leipzig.de/en/download/Guarani} and 2000 from another Guaraní corpus \cite{chiruzzo-etal-2020-development} into Spanish pronunciation and ran them through a Spanish TTS system. In the case of Italian, we pulled sentences from unused pairs in the Mozilla data we had downloaded for training and ran them through Arabic and Spanish transliteration engines and TTS.  Details for all the datasets we worked with are in Table \ref{tab_data}. Our data and software  will be available upon acceptance.

% \nrr{Drafting}

% We used Microsoft TTS\footnote{https://azure.microsoft.com/en-us/services/cognitive-services/text-to-speech/} to generate synthetic data using Arabic and Spanish TTs systems, but with orthographic transfer.

% We also used Microsoft Kiswahili TTS to create synthetic Kiswahili corpus. Using synthetic data created from a TTS of the same language can help in situations where the available data is privately owned such as the case of Kiswahili. Table \ref{tab_data} gives a breakdown of our corpora.

\begin{table}
\centering
\scalebox{0.80}{
\small\addtolength{\tabcolsep}{-4pt}
\begin{tabular}{lcccccc}
\hline
~ & \textit{Utter-} & \textit{train/val/} &  ~ & \textit{Spea-} & \textit{Gen-} & \textit{TTS} \\
\textit{Lang. + Corpus} & \textit{ances} & \textit{test split} & \textit{Hours} & \textit{kers} & \textit{der} & \textit{API} \\

\hline
\textit{\textbf{Kiswahili}}\\
~~~Gamayun SWH & 4700 & 3900/400/400 & 6.06 & 1 & M  & N/A \\
% ~~~~~~validation and test sets each consist of 400 utterances from this set \\
% ~~~Gamayun SWH (dev) & 400 & ~ & 1 & M  & N/A \\
% ~~~Gamayun SWH (test) & 400 & ~ & 1 & M  & N/A \\
~~~SWH-TTS & 8465 & 8465/0/0 & ~14.5 & 4 & M/F & Microsoft \\
~~~ARA-TTS & 14737 & 14737/0/0 & ~27.7  & 4 & M/F & Google\\
~~~ITA-TTS & 14737 & 14737/0/0 & ~26.2 & 4 & M/F & Google \\
~~~SPA-TTS & 8000 & 8000/0/0 & 12.6 & 3 & M/F & Google \\
~~~TUR-TTS & 8000 & 8000/0/0 & 13.1 & 5 & M/F & Google\\
~~~ARA-TTS-trans & 4000 & 4000/0/0 & 8.13  & 2 & M/F & Microsoft\\
~~~SPA-TTS-trans & 3008 & 3008/0/0 & 4.58  & 2 & M/F & Microsoft\\

\hline
\textit{\textbf{Guaraní}}\\
~~~Mozilla GRN & 1883 & 1083/400/400 & 2.31 & 55 & M/F & N/A \\
% ~~~SPA-TTS-trans & 4000 & 8.3 & 2 & M/F & Microsoft \\
~~~AFR-TTS & 8000 & 8000/0/0 & 19.0 & 1 & F & Google \\
~~~SPA-TTS & 8000 & 8000/0/0 & 15.5 & 3 & M/F & Google \\
~~~FRA-TTS & 8000 & 8000/0/0 & 17.6 & 5 & M/F & Google \\

\hline
\textit{\textbf{Suba}}\\
~~~African Voices & 1178 & 942/118/118 & 1.7 & 1 & M & N/A \\
% ~~~African Voices (dev) & 1178 & 1.7 & 1 & M & N/A \\
% ~~~African Voices (test) & 1178 & 1.7 & 1 & M & N/A \\
~~~SPA-TTS & 7536 & 7536/0/0 & 14.28 & 4 & M/F & Google \\
~~~ARA-TTS & 7536 & 7536/0/0 & 18.92 & 4 & M/F & Google \\
% ~~~ Bible.is test & 163 & 0.37 &  1 & M & N/A \\

\hline
\textit{\textbf{Italian}}\\
~~~Mozilla ITA & 4700 & 3900/400/400 & 8.23 & 1562 & M/F & N/A \\ 
~~~RON-TTS & 8000 & 8000/0/0 & 10.7 & 1 & F & Google \\
~~~SPA-TTS & 8000 & 8000/0/0 & 8.60 & 3 & M/F & Google \\
~~~FIN-TTS & 8000 & 8000/0/0 & 9.73 & 1 & F & Google \\
% ~~~dev &  &  &  & M/F  & N/A \\ 
% ~~~test &  &  & & M/F  & N/A \\ 

\hline
\textit{\textbf{Indonesian}} \\
~~~Mozilla IND & 4000 & 4000/0/0 & 4.52 & 273 & M/F & N/A\\

\hline
\end{tabular}
}
\caption{Speech corpora details. TTS-trans refers to TTS audio from transliterated text. Note that in many experiments we restrict training sets further. Data with train/val/test splits and details about subsets for each experiment are available at \url{https://github.com/n8rob/Multilingual_TTS_Augmentation}.
}
\label{tab_data}
\vspace{-10mm}
\end{table}

% \footnote{We used the Arabic-Jordan setting on Microsoft Azure, but the synthesized speech is actually Modern Standard Arabic, not Jordanian Colloquial Arabic.}

% \subsection{ASR Experiments}
% %\quad
% We conducted three categories of experiments:

% \begin{enumerate}
% \item Kiswahili ASR with a comparison of different augmentation methods (Tables \ref{tab:og_results_lrec} and \ref{tab:og_results_hrec})
% % We split the authentic training set into 2 sets; 300 and 3900 sentences. We also selected three sets from the synthetic TTS data; 300, 4000 and 14373 sentences.

% \item Guaraní ASR with Spanish TTS augmentation, to illustrate a lower-resourced setting (Table \ref{Guaraní_results})

% \item Italian ASR with TTS augmentation from a similar language, Spanish, and a distant language, Arabic, to explore the effect of language similarity (Table \ref{tab:ito_rez})
% % We had two sets of authentic training set like above. The synthetic data from TTS systems was  divided into two; 300 and 4000 sentences.

% \end{enumerate} 

\subsection{Experimental Setup}
We combine different amounts of TTS data and authentic data  to train an ASR system.  We evaluate and test on authentic data only. 
We used ESPNet2 \cite{Watanabe2018} with the default RNNTransducer \cite{9688251} and
RNNLM to train our End-to-End ASR systems. Figure \ref{pipeline} is a visual depiction of our approach. 

\section{Results}
\label{sxn:res}

\begin{figure}[h!]
    \centering
    \includegraphics[width=8cm]{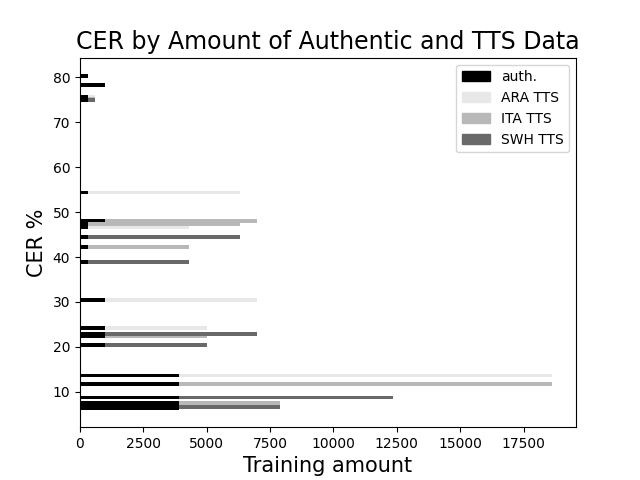}
\caption{\label{fig:amt}
        \centering
    %The three dots represent beginning with 300, 1000, and 3900 authentic data pairs. 
    Different-shaded bars represent data of varying number of utterances from authentic/synthetic sources for Kiswahili. Note that for each amount  of authentic data (black bars), adding 4000 pairs of synthetic data decreases error rate, but that improvement lessens with more synthetic data.
    % In the lower-resource settings, TTS augmentation reduces error rates but eventually saturates due to over-fitting on synthetic data
    }
\end{figure}

\begin{figure}[h!]
    \centering
    \includegraphics[width=8cm]{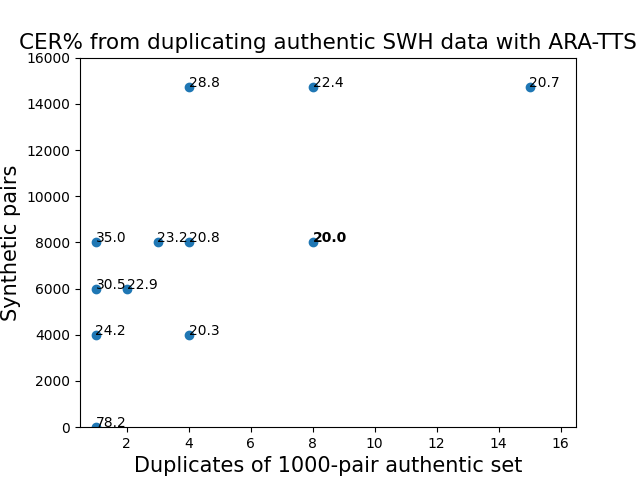}
\caption{\label{fig:duplicate}
        \centering
    Character error rate (CER) values from duplicating authentic 1000-pair Kiswahili set alongside increased synthetic data from an Arabic TTS system. Note that dots along the diagonal with positive slope represent balanced data and tend to perform better. The optimal approach here is using 8000 synthetic pairs and copying the authentic data 8 times. %(to balance).
    }
\end{figure}

We explored effectiveness of TTS augmentation through a pivot language on three axes: synthetic data amount given authentic data, choice of pivot language, and TTS audio quality. 

\subsection{Synthetic data amount} 
\label{sxn:synth_data_amt}

We began our experiments by probing for the optimal amount of TTS augmented data. For these experiments we augmented Kiswahili speech data with Arabic, Italian, and Kiswahili TTS. We chose Arabic because of its historical influence on Kiswahili and Italian because its output for Kiswahili text sounded reasonably accurate to a proficient Kiswahili speaker. %We chose Arabic because of its historical influence on Kiswahili and Italian because its TTS system produced reasonable Kiswahili speech, according to a proficient speaker.
% Throughout tables and figures, we use ISO 639-3 codes to refer to the languages mentioned: AFR (Afrikaans), ARA (Arabic), FIN (Finnish), FRA (French), GRN (Guaraní), ITA (Italian), RON (Romanian), SPA (Spanish), SXB (Suba), SWH (Kiswahili), TUR (Turkish)}.
Results are in Figures \ref{fig:amt} and \ref{fig:duplicate}. Because low-resource languages can have varying amounts of transcribed audio data, we ran experiments using three authentic training sets of sizes 300, 1000, and 3900. Our test set of 400 utterances was kept constant throughout the experiments. In each of these settings, increasing the amount of TTS synthetic data improves error rates until it reaches a point where further increase degrades performance. This degradation occurs when the model overfits on the synthetic data and thus performs poorly when tested on authentic data. Figure \ref{fig:amt} illustrates that in the case of each authentic data amount, using $4000$ augmented pairs improves performance (or stagnates in the highest-resource case), and continuing to add synthetic data beyond that degrades performance.

To avoid this data imbalance and over-fitting, we duplicated authentic data in the setting with 1000 authentic pairs. Results are in Figure \ref{fig:duplicate}. We found that when beginning with 1000 transcribed recordings, duplicating authentic data eight times and augmenting with 8000 TTS synthetic pairs works best. This is a relevant example since many low-resource languages have roughly 1000 transcribed recordings, or 1-2 hours of speech, available. (E.g. our Guaraní and Suba data, see Table \ref{tab_data}.) As shown in Figure \ref{fig:duplicate}, this policy reduces CER from $78.2\%$ (no duplication or augmentation) to $20.0\%$ for $74.4\%$ CERR for Kiswahili with an Arabic TTS system. It works well for other language combinations as well: $31.4\%$ CERR for Italian (with Finnish TTS), $64.5\%$ CERR for Guaraní (with French TTS), and $45.0\%$ CERR for Suba (with Spanish TTS).

\subsection{Choice of pivot language} 
\label{sxn:pivot_lang}

Before conducting explorations related to pivot language choice, we analyzed the textual outputs from the lowest- and highest-resource experiments featured in Figure \ref{fig:amt} beyond error rate. We tested whether there is a relationship between the pivot language or amount and the characters that the system learns to recognize. We did not find any noticeable pattern for phones that was tied to TTS augmentation. The only system that recognized any characters at a rate of $\geq25\%$ outside the average was the Kiswahili system with 3900 authentic pairs and 4000 Arabic TTS pairs, on $5$ of the $26$ Kiswahili characters.\footnote{The average described here is the average proportion of substitutions and deletions that occurred for each target language character. The five characters at least 25\% outside the norm for one of the data configurations were \textit{s, u, i, l, j}.}%\footnote{Over the configurations in Figure \ref{fig:amt} using 300 and 3900 authentic pairs, we averaged the proportions of substitutions and deletions that occurred for each target language character. The only cases where any of the proportions were greater than $1.25$ times the average (or where the average was greater than $1.25$ times the proportion in question) were the letters \textit{s, u, i, l, j} for the system with 3900 authentic pairs and 4000 Arabic TTS pairs. (See Figure \ref{fig:amt}.)} %\nrr{Not sure if we need this paragraph} \drm{When in doubt, leave it out.}

We experimented to test whether choosing a pivot language with high relatedness to the target language can improve results. We tested on three ASR target languages: Italian, Kiswahili, and Guaraní. Results are in Table \ref{tab:langsim}. We used pre-computed \texttt{lang2vec} distances from URIEL \cite{littell2017uriel} to determine language relatedness. Following best practice recommendations \cite{wu2021cross}, we relied primarily on geographical and genetic distance. We restricted our similarity search to the languages supported by Google Cloud TTS, and we added some language pairs to test other language characteristics.\footnote{Google Cloud TTS does not support any languages geographically or genetically close to Guaraní or Kiswahili other than Spanish and Arabic, respectively. We selected Italian, Spanish, and Turkish (TUR) for Kiswahili to explore if their straightforward orthography systems would yield an advantage. For Italian, we chose  Romanian as they are close geographically and genetically per URIEL and Spanish because of their linguistic proximity. We chose Afrikaans (AFR) for Guaraní since URIEL finds them phonologically close. We also tested languages that are closest to the \textit{average} geographical/genetic distance from the target language: Finnish (FIN) for Italian and French (FRA) for Guaraní. } %For each target language we found a close pivot language and the pivot language with the closest-to-average distance from the target, out ofx the candidate set of all the languages supported by Google Cloud TTS. Because Google Cloud does not support any languages that are genetically or geographically close to Guaraní in origin, we tested with Afrikaans, which was the closest to Guaraní phonologically, per URIEL.

In our experiments featured in Table \ref{tab:langsim}, language similarity did not appear to determine TTS augmentation suitability. The best-performing TTS system for Italian, for example, was Finnish, and for Guaraní, French. Arabic performed best for Kiswahili, but this is likely because the data amount configuration was tuned for this pair. (Italian TTS also fares well.) It is possible that having a diversity of language characteristics provided by the TTS system is actually an advantage. We conclude that for any ASR target language, multiple TTS pivot languages should be tried to determine one that works well.

\begin{table}[h!]
\centering
\begin{tabular}{cccc}
\hline
 \textit{Target} & \textit{Pivot} & & \\
\textit{lang.} & \textit{lang.} & \textit{WER} & \textit{CER} \\
\hline
ITA & RON & 98.8\% & 73.1\% \\
ITA & SPA & \textbf{91.7\%} & 53.9\% \\
ITA & FIN & 92.5\% & \textbf{53.5\%} \\
\hline
SWH & ARA & \textbf{51.6\%} & \textbf{20.0\%} \\
% SWH & YUE (avg) & -\% & -\% \\
SWH & ITA & 53.3\% & 24.3\% \\
SWH & SPA & 62.8\% & 22.7\% \\
SWH & TUR & 69.3\% & 29.9\% \\
\hline
GRN & AFR & 79.6\% & 34.4\% \\
GRN & SPA & 74.1\% & \textbf{30.0\%} \\
GRN & FRA & \textbf{73.0\%} & \textbf{30.0\%} \\
% ARA & 1000$\times$8/8000 & \textbf{51.6\%} & \textbf{20.0\%} \\
% ARA & 1000$\times$4/14737 & 61.2\% & 28.8\% \\
% ARA & 1000$\times$8/14737 & 53.2\% & 22.4\% \\
% ARA & 1000$\times$15/14737 & \textbf{51.6\%} & 20.7\% \\
\hline
\end{tabular}
\caption{\label{tab:langsim}
\textbf{Language choice results.} All of these models were trained with 1000 authentic pairs copied eight times and 8000 synthetic TTS pairs, the optimal amounts for starting with 1000 authentic pairs, per our experiments in Figure \ref{fig:duplicate}. All synthetic audio was generated by Google Cloud with standard settings. Best results are \textbf{bold}; note that the SWH-ARA combination was tuned for this data configuration.
}
\vspace{-8mm}
\end{table}
\subsection{TTS quality} 
\label{sxn:tts_qual}

We sought to improve TTS quality in order to improve augmentation and ASR performance via (1) using Microsoft TTS rather than Google Cloud (higher quality but more time consuming), and (2) transliterating TTS text into pivot language orthography, as discussed in \S \ref{sxn:data_method}. %\pao{Nate cite Epitran so that we resolve this cyclic situation} 
This technique seems in fact to increase TTS quality: in an A-B test on 20 recording comparisons between Kiswahili transliterated and not transliterated for Arabic TTS, transliterated audio was preferred 100\% of the time by a proficient Kiswahili speaker. Surprisingly, higher quality of TTS augmentation data does not always aid ASR performance. See Table \ref{tab:ttsqual}, where we performed these experiments on Kiswahili ASR with Arabic and Spanish TTS augmentation. Using transliterated Spanish TTS tends to have an advantage over lower-quality Spanish TTS, but higher quality Arabic TTS actually hurts performance consistently across starting amounts. 

A blind TTS quality test by a proficient Kiswahili speaker on 20 utterance comparisons gave the following quality scores: \textit{poor ARA}:$0$, \textit{trans ARA}:$4$, \textit{poor SPA}:$4$, \textit{trans SPA}:$12$. This is significant because although the Spanish TTS is clearly higher quality, it performs worse for ASR in the optimal settings in Table \ref{tab:langsim}. 
Similarly, transliterated Arabic which is of higher quality actually hurts performance as compared to un-transliterated Arabic TTS. Spanish and Arabic scores in Table \ref{tab:ttsqual} are not directly comparable because of the difference in training set sizes. 

\begin{table}[h!]
\centering
\begin{tabular}{lccccc}
\hline
 \textit{Target} & \textit{Auth.} & \textit{Aug.} & & \\
\textit{lang.} & \textit{pairs} & \textit{data} & \textit{WER} & \textit{CER} \\
\hline
SWH & 300 & poor ARA & 83.1\% & 46.6\% \\
SWH & 300 & trans ARA & 87.7\% & 53.2\% \\
SWH & 300 & poor SPA & 93.6\% & \textbf{43.8\%}  \\
SWH & 300 & trans SPA & \textbf{82.2\%} & 45.1\% \\
\hline
SWH & 1000 & poor ARA & 56.6\% & 24.2\% \\
SWH & 1000 & trans ARA & 61.2\% & 27.1\% \\
SWH & 1000 & poor SPA & 77.4\% & 32.9\% \\
SWH & 1000 & trans SPA & \textbf{54.6\%} & \textbf{21.7\%} \\
\hline
SWH & 3900 & poor ARA & 22.7\% & 7.7\% \\
SWH & 3900 & trans ARA & 23.0\% & 7.8\% \\
SWH & 3900 & poor SPA & 24.9\% & 8.0\% \\
SWH & 3900 & trans SPA & \textbf{20.5\%} & \textbf{6.5\%} \\
\hline
\end{tabular}
\caption{\label{tab:ttsqual}
\textbf{TTS quality experiments.} %A-B tests were performed by a proficient speaker of the target language. 
Models were trained using 4000 Arabic TTS synthetic pairs or 3008 Spanish; no authentic data duplication. %Arabic and Spanish scores cannot be directly compared because of the size difference. 
 %\nrr{We may want to re-run these with the optimal data amount of eight times authentic plus 8000 TTS. We may also want to run them all on another language, in addition to Kiswahili with Arabic. Thoughts?}
% \nrr{Add ABCD test}
}
\vspace{-8mm}
\end{table}

We ran an ablation experiment with no authentic Kiswahili data: we trained solely on 4000 Arabic TTS files, both transliterated and not, with the same authentic test and validation sets. The higher-quality transliterated TTS performed slightly worse, with \textit{WER=102.5\%} and \textit{CER=72.9\%} compared to \textit{WER=99.6\%} and \textit{CER=72.7\%} for the lower-quality set.

These findings prompted questions as to how much TTS accuracy matters in ASR training. As another ablation, we augmented training data for Kiswahili ASR with 4000 Indonesian (IND) speech files, not resembling Kiswahili at all. Interestingly, in the setting with 3900 authentic pairs, this had a similar effect to using TTS and outscored multiple TTS examples (with \textit{CER=7.2\%} compared to \textit{CER=7.8\%} for ARA-TTS augmentation of the same amount). In the setting of 300 authentic pairs, augmentation quality is more relevant: Indonesian noise does not improve error rates significantly, but TTS audio does.

\subsection{Application to low-resource languages}

\begin{table}[h!]
\centering
\begin{tabular}{lcccc}
\hline
 & \textit{Augmentation} & \textit{WER} & \textit{CER} \\
\hline
\textbf{GRN:} & no aug. & 100\% & 84.4\% \\
 & FRA TTS & \textbf{73.0\%} & \textbf{30.0\%} \\

\hline
\textbf{SXB:} & no aug. & 86.3\% & 51.6\% \\
 & SPA TTS & \textbf{69.7\%} & \textbf{28.4\%} \\
 
\hline
\end{tabular}
\caption{\label{Guaraní_results}
\textbf{We achieve significant error rate reductions in low-resource Guaraní and Suba ASR by applying findings.}
}
\label{tab:guarani}
\vspace{-6mm}
\end{table}

% \nrr{This is where we put our results for Guaraní and Suba (and Kiswahili?) by employing all the strategies we explored. This is the section of experiments where we'll get our x\% and y\% mentioned in the abstract and introduction.}

% \nrr{Essentially we'll want to use roughly 8000 TTS pairs and duplicate authentic data to balance that. We won't worry about TTS quality since it doesn't seem to matter. We will search over a few TTS pivot languages (or perhaps just use one that seems to work well for African languages or universally).}

% \nrr{Dear Nate please do this}

% Low-resource Guaraní ASR benefits from the improvement of 4000 synthetic pairs. (See Table \ref{Guaraní_results}.) In this case, performance can improve dramatically and be on par with higher-resource experimental results. Our results also suggest that in higher-resourced settings, TTS augmentation is less effective, and TTS quality and language do not effect performance and that the size of synthetic data should be carefully chosen to avoid overfitting.  %\nrr{We need to actually perform more experiments here, i.e. domain congruence etc.}

For many low-resource languages, only a small set of transcribed audio is available: on the order of 1000 utterances or 1-2 hours. Two such datasets are the CommonVoice Guaraní and African Voices Suba datasets. 
Employing our findings in practice, we augmented both of these datasets by duplicating the authentic data eight times and adding an equal amount of TTS synthetic data, following the recommended procedure based on Figure \ref{fig:duplicate}. See Table \ref{tab:guarani}. This augmentation results in better ASR, with word error reduction rates (WERR) of $27.0\%$ and $19.2\%$ and CERR $64.5\%$ and $45.0\%$.

\section{Conclusion}
% \nrr{Needs updating}

We show that synthetic audio from a high-resource pivot language TTS system can be used to augment authentic datasets and improve ASR for low-resource languages. Our experiments suggest that performance improves best when several thousand TTS-generated synthetic pairs are used and authentic data is replicated to an equal amount, and when a search over potential pivot languages is conducted. Our experiments suggest, surprisingly, that measured TTS audio quality may not effect suitability for ASR training augmentation. These techniques improve ASR for low-resource languages Kiswahili, Guaraní, and Suba ($74.4\%$, $64.5\%$, and $45.0\%$ CERR, respectively). They are also broadly applicable to the thousands of other low-resource languages often overlooked in speech technologies. This has promising implications of increased information access for speakers of these languages and for documentation and revitalization efforts for endangered languages like Suba.

Future work may involve methodological advances, such as including authentic speech from the TTS pivot language in training; or pretraining steps, such as training a model on noisy TTS synthetic data and then tuning on authentic data. This process could theoretically be enhanced by using pre-trained representations from multilingual self-supervised models such as XLSR \cite{conneau21_interspeech}. Further investigation may also involve more rigorous analyses of optimal pivot languages, including comparisons of language recording acoustic features. %Future researchers may also seek to apply these techniques to a larger and more diverse group of endangered languages.

% \nrr{TO-do list:
% \begin{itemize}
%     \item Emphasize 14.4 percent in Table \ref{tab:allres}
%     \item Improve this score by generating more TTS (until saturation)
%     \item Copy authentic data to avoid TTS saturation and overfitting
%     \item Replicate experiments for Guaraní case
%     \item Tune ESPNet parameters to improve ASR
%     \item Access loss plots and other information in ESPNet
%     \item Other ideas?
% \end{itemize}
% }

%Future work in this area includes finding exact thresholds for optimal TTS synthetic data amounts, exploring how the traditional methods of ASR data compare with using Synthetic TTS data for particular languages. More experiments can be done with synthetic data that is not generated with text from the target language and also using a language which is farther in phones from the target language like English. Further, more work can be done to explore the effects of TTS quality on ASR performance. Since the transliterated TTS did not improve significantly over noisy TTS, it is possible that TTS quality or even language may not have an effect on its effectiveness.

\section{Acknowledgements}

We thank Brian Yan, Alan W Black, Xinjian Li, Graham Neubig, and Rebeca Knapp for their contributions and support.

\clearpage

\bibliographystyle{IEEEtran}

\bibliography{mybib}

% \begin{thebibliography}{9}
% \bibitem[1]{Davis80-COP}
%   S.\ B.\ Davis and P.\ Mermelstein,
%   ``Comparison of parametric representation for monosyllabic word recognition in continuously spoken sentences,''
%   \textit{IEEE Transactions on Acoustics, Speech and Signal Processing}, vol.~28, no.~4, pp.~357--366, 1980.
% \bibitem[2]{Rabiner89-ATO}
%   L.\ R.\ Rabiner,
%   ``A tutorial on hidden Markov models and selected applications in speech recognition,''
%   \textit{Proceedings of the IEEE}, vol.~77, no.~2, pp.~257-286, 1989.
% \bibitem[3]{Hastie09-TEO}
%   T.\ Hastie, R.\ Tibshirani, and J.\ Friedman,
%   \textit{The Elements of Statistical Learning -- Data Mining, Inference, and Prediction}.
%   New York: Springer, 2009.
% \bibitem[4]{YourName17-XXX}
%   F.\ Lastname1, F.\ Lastname2, and F.\ Lastname3,
%   ``Title of your INTERSPEECH 2022 publication,''
%   in \textit{Interspeech 2022 -- 23\textsuperscript{rd} Annual Conference of the International Speech Communication Association, September 18-22, Incheon, Korea, Proceedings, Proceedings}, 2022, pp.~100--104.
% \end{thebibliography}

\end{document}